\documentclass[runningheads]{llncs}

\usepackage{palatino,epsfig,latexsym,natbib}
\usepackage{amsmath,amssymb}
\usepackage{bm}
\usepackage{hyperref}
\usepackage{subcaption}
\usepackage{color}
\usepackage{graphicx}
\usepackage{rotating}
\usepackage{booktabs}
\usepackage[ruled]{algorithm2e}
\usepackage{setspace}
\usepackage{multirow}
\usepackage{xcolor,colortbl}
\colorlet{transpblue}{white!50!blue}
\newcolumntype{a}{>{\columncolor{transpblue}}l}

\begin{document}

\title{GENDIS: GENetic DIscovery of Shapelets}  

\author{Gilles Vandewiele \and Femke Ongenae \and Filip De Turck
}

\institute{{IDLab, Ghent University -- imec, 9052 Ghent, Belgium
	\email{\{firstname\}.\{lastname\}@ugent.be}}}

\maketitle

\begin{abstract}

In the time series classification domain, shapelets are small time series that are discriminative for a certain class. It has been shown that classifiers are able to achieve state-of-the-art results on a plethora of datasets by taking as input distances from the input time series to different discriminative shapelets. Additionally, these shapelets can easily be visualized and thus possess an interpretable characteristic, making them very appealing in critical domains, such as the health care domain, where longitudinal data is ubiquitous. In this study, a new paradigm for shapelet discovery is proposed, which is based upon evolutionary computation. The advantages of the proposed approach are that (i) it is gradient-free, which could allow to escape from local optima more easily and to find suited candidates more easily and supports non-differentiable objectives, (ii) no brute-force search is required, which drastically reduces the computational complexity by several orders of magnitude, (iii) the total amount of shapelets and length of each of these shapelets are evolved jointly with the shapelets themselves, alleviating the need to specify this beforehand, (iv) entire sets are evaluated at once as opposed to single shapelets, which results in smaller final sets with less similar shapelets that result in similar predictive performances, and (v) discovered shapelets do not need to be a subsequence of the input time series. We present the results of experiments which validate the enumerated advantages.

\end{abstract}


\section{Introduction} \label{sec:intro}

\subsection{Background}
Due to the uprise of the Internet-of-Things (IoT), mass adoption of sensors in all domains, including critical domains such as health care, can be noticed. These sensors produce data of a longitudinal form, i.e. time series. Time series differ from classical tabular data, since a temporal dependency is present where each value in the time series correlates with its neighboring values. One important task that emerges from this type of data is the classification of time series in their entirety. A model able to solve such a task can be applied for a wide variety of applications, such as distinguishing between normal brain activity and epileptic activity~\citep{chaovalitwongse2006electroencephalogram}, determining different types of physical activity~\citep{liu2015sensor}, or profiling electronic appliance usage in smart homes~\citep{li2016profiling}. Often, the largest discriminative power can be found in smaller subsequences of these time series, called shapelets. Shapelets semantically represent intelligence on how to discriminate between the different targets of a time series dataset~\citep{grabocka2014learning}. We can use a set of shapelets and the corresponding distances from each of these shapelets to each of the input time series as features for a classifier. It has been shown that such an approach outperforms a nearest neighbor search based on dynamic time warping distance on almost every dataset, which was deemed to be the state-of-the-art for a long time~\citep{abanda2019review}. Moreover, shapelets possess an interpretable characteristic since they can easily be visualized and be retraced back to the input signal, making them very interesting for decision support applications in critical domains, such as the medical domain. In these critical domains, it is of vital importance that a corresponding explanation can be provided alongside a prediction, since a wrong decision can have a significant negative impact.

\subsection{Related Work}\label{sec:related_work}
Shapelet discovery was initially proposed by~\cite{ye2009time}. Unfortunately, the initial algorithm quickly becomes intractable, even for smaller datasets, because of its large computational complexity ($O(N^2M^4)$, with $N$ the number of time series and $M$ the length of the smallest time series in the dataset). This complexity was improved two years later, when~\cite{mueen2011logical} proposed an extension to this algorithm, that makes use of caching for faster distance computation and a better upper bound for candidate pruning. These improvements reduce the complexity to $O(N^2M^3)$, but have a larger memory footprint. \cite{rakthanmanon2013fast} proposed an approximative algorithm, called Fast Shapelets (\textsc{\texttt{fs}}), that finds a suboptimal shapelet in $O(NM^2)$ by first transforming the time series in the original set to Symbolic Aggregate approXimation~(\textsc{sax}) representations~\citep{lin2003symbolic}. Although no guarantee can be made that the discovered shapelet is the one that maximizes a pre-defined metric, they show they are able to achieve very similar classification performances, empirically on 32 datasets. \\

All the aforementioned techniques search for a single shapelet that optimizes a certain metric, such as information gain. Often, one shapelet is not enough to achieve good predictive performances, especially for multi-class classification problems. Therefore, the shapelet discovery is applied in a recursive fashion in order to construct a decision tree. \cite{lines2012shapelet} proposed Shapelet Transform (\textsc{\texttt{st}}), which performs only a single pass through the time series dataset and maintains an ordered list of shapelet candidates, ranked by a metric, and then finally takes the top-k from this list in order to construct features. While the algorithm only performs a single pass, the computational complexity still remains to be $\mathcal{O}(N^2M^4)$, which makes the technique intractable for larger datasets. Extensions to this technique have been proposed in the subsequent years which drastically improved the performance of the technique~\citep{hills2014classification,bostrom2017binary}. \cite{lines2012shapelet} compared their technique to 36 other algorithms for time series classification on 85 datasets~\citep{bagnall2016great}, which showed that their technique is one of the top-performing algorithms for time series classification and the best-performing shapelet extraction technique in terms of predictive performance.\\

\cite{grabocka2014learning} proposed a technique where shapelets are learned through gradient descent, in which the linear separability of the classes after transformation to the distance space is optimized, called Learning Time Series Shapelets~(\textsc{\texttt{lts}}). The technique is competitive to \textsc{\texttt{st}}, while not requiring a brute-force search, making it tractable for larger datasets. Unfortunately, \textsc{\texttt{lts}} requires the user to specify the number of shapelets and the length of each of these shapelets, which can result in a rather time-intensive hyper-parameter tuning process in order to achieve a good predictive performance. Three extensions of \texttt{\textsc{lts}} which improve the computational runtime of the algorithm, have been proposed in the subsequent years. Unfortunately, in order to achieve these speedups, predictive performance had to be sacrificed. A first extension is called Scalable Discovery (\texttt{\textsc{sd}})~\citep{grabocka2015scalable}. It is the fastest of the three extensions, improving the runtime two to three orders of magnitude, but at a cost of having a worse predictive performance than \texttt{\textsc{lts}} on almost every tested dataset. Second, in 2015, Ultra-Fast Shapelets (\texttt{\textsc{ufs}})~\citep{wistuba2015ultra} was proposed. It is a better compromise of runtime and predictive performance, as it is an order of magnitude slower than \texttt{\textsc{sd}}, but sacrifices less of its predictive performance. A final and most recent extension is called Fused LAsso Generalized eigenvector method (\texttt{\textsc{flag}})~\citep{hou2016efficient}. It is the most notable of the three extensions as it has runtimes competitive to \texttt{\textsc{sd}} while being only slightly worse than \texttt{\textsc{lts}} in terms of predictive performance. 

\subsection{Our Contribution}
In this paper, we introduce an evolutionary algorithm, \textsc{\texttt{gendis}}, that discovers a set of shapelets from a collection of labeled time series. The aim of the proposed algorithm is to achieve state-of-the-art predictive performances similar to the best-performing algorithm, \textsc{\texttt{st}}, with a smaller number of shapelets, while having a low computational complexity similar to \textsc{\texttt{lts}}. \\

The goal of \textsc{\texttt{gendis}} is to retain some of the positive properties from \textsc{\texttt{lts}} such as its scalable computational complexity, the fact that entire sets of shapelets are discovered as opposed to single shapelets and that it can discover shapelets outside the original dataset. We demonstrate the added value of these two final properties through intuitive experiments in Subsections~\ref{subsec:sets} and \ref{subsec:outside} respectively. Moreover, \textsc{\texttt{gendis}} has some benefits over \textsc{\texttt{lts}}. First, genetic algorithms are gradient-free, allowing for any objective function and an easier escape from local optima. Second, the total amount of shapelets and the length of each of these shapelets do not need to be defined prior to the discovery, alleviating the need to tune this, which could be computationally expensive and may require domain knowledge. Finally, we show by a thorough comparison, in Subsection~\ref{subsec:comparison}, that \textsc{\texttt{gendis}} empirically outperforms \textsc{\texttt{lts}} in terms of predictive performance.

\section{Methodology}\label{sec:methodology}

In the following section we will first explain some general concepts from the time series analysis and shapelet discovery domain, on which we will then build further to elaborate our proposed algorithm, \textsc{\texttt{gendis}}.

\subsection{Time series Matrix \& Label Vector}\label{subsec:method_data}
The input to a shapelet discovery algorithm is a collection of $N$ time series. For ease of notation, we will assume that the time series are synchronized and have a fixed length of $M$, resulting in an input matrix $\bm{T} \in \mathbb{R}^{N \times M}$. It is important to note that \textsc{\texttt{gendis}} could perfectly work with variable length time series as well. In that case, $M$ would be equal to the minimal time series length in the collection. Since shapelet discovery is a supervised approach, we also require a label vector $\bm{y}$ of length $N$, with each element $y_i \in \{1, \ldots, C\}$ and $C$ the number of classes and $y_i$ corresponding to the label of the $i$-th time series in $\bm{T}$.

\subsection{Shapelets and Shapelet Sets}\label{subsec:method_shapelet}
Shapelets are small time series which semantically represent intelligence on how to discriminate between the different targets of a time series dataset. In other words, they are very similar to subsequences from time series of certain (groups of) classes, while being dissimilar to subsequences of time series of other classes. The output of a shapelet discovery algorithm is a collection of $K$ shapelets, $\bm{S}~=~\{s_1, \ldots, s_K\}$, called a shapelet set. In \textsc{\texttt{gendis}}, $K$ and the length of each shapelet does not need to be set beforehand and each shapelet can have a variable length, smaller than $M$. These $K$ shapelets can then be used to extract features for the time series, as we will explain subsequently.

\subsection{Distance Matrix Calculation}\label{subsec:method_distance}
Given an input matrix $\bm{T}$ and a shapelet set $\bm{S}$, we can construct a pairwise distance matrix $\bm{D} \in \mathbb{R}^{N \times K}$:

\begin{equation*}
dist(\bm{S}, \bm{T}) = \bm{D}
\end{equation*}

The distance matrix, $\bm{D}$, is constructed by calculating the distance between each ($t$, $s$)-pair, where $t \in \bm{T}$ is an input time series and $s \in \bm{S}$ a shapelet from the candidate shapelet set. This matrix can then be fed to a machine learning classifier. Often, $K~<<~M$ such that we effectively reduce the dimension of our data. In order to calculate the distance from a shapelet $s$ in $\bm{S}$ to a time series $t$ from $\bm{T}$, we convolute the shapelet across the time series and take the minimum distance:
\begin{equation*}
dist(s, t) = \min_{1\ \leq\ i\ \leq\ |t|-|s|} d(s, t[i:i+|s|-1])
\end{equation*}
with $d(.)$ a distance metric, such as the Euclidean distance, and $t[i:i+|s|-1]$ a slice from $t$ starting at index $i$ and having the same length as $s$.

\subsection{Shapelet Set Discovery Objective}\label{subsec:shapelet_extraction}
Conceptually, the goal of a shapelet set extraction technique is to find a set of shapelets, $\bm{S}$, that produces a distance matrix, $\bm{D}$, that minimizes the loss function, $\mathcal{L}$, of the machine learning technique to which it is fed, $h(.)$, given the ground truth, $\bm{y}$. 

\begin{equation*}
\min_{\bm{S}} \mathcal{L}(h(dist(\bm{T}, \bm{S})), \bm{y})
\end{equation*} \\

\begin{figure*}
	\centering
	\includegraphics[width=1.0\textwidth]{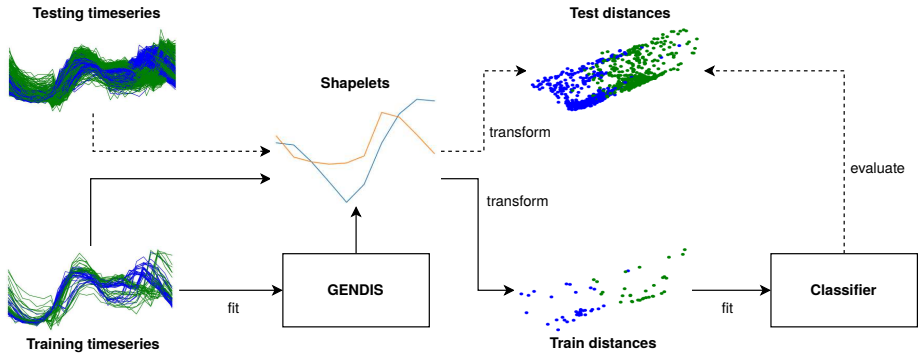}
	\caption{A schematic overview of shapelet discovery. First, shapelets are independently extracted using the training set. These shapelets are then used to transform the train and test set in features. A classifier can afterwards be fit on the training features and evaluated on the test features.}
	\label{fig:gendis_overview}      
\end{figure*}

It should be noted that the shapelet extraction and classification phases are completely decoupled, as depicted in Figure~\ref{fig:gendis_overview}. A set of shapelets ($\bm{S}$) is first independently mined, which is then used to transform the time series into features that correspond to distances from each of the time series to the shapelets in the set. These features are then fed to a machine learning classifier.

\subsection{\textsc{\texttt{gendis:}} GENetic Discovery of Interpretable Shapelets} \label{subsec:genetic}

In this paper, we propose a genetic algorithm that evolves a set of variable-length shapelets, $\bm{S}$, in $\mathcal{O}(NM^2)$ which produces a distance matrix $\bm{D}$, based on a collection of time series $\bm{T}$, that results in optimal predictive performance when provided to a machine learning classifier. The intuition behind the approach is similar to \textsc{\texttt{lts}}, which we mentioned in Section~\ref{sec:related_work}, but the advantage is that both the size of $\bm{S}$, $K$, and the length of each shapelet $s \in \bm{S}$ are evolved jointly, alleviating the need to specify the number of shapelets and the length of each shapelet prior to the extraction. Moreover, the technique is gradient-free, which allows for non-differentiable objectives and to escape local optima more easily. \\

The building blocks of a genetic algorithm consist of at least a crossover, mutation and selection operator~\citep{mitchell1998introduction}. Additionally, we seed, or initialize, the algorithm with specific candidates instead of completely random candidates~\citep{julstrom1994seeding} and apply elitism~\citep{sheble1995refined} to make sure the fittest candidate set is never discarded from the population or never experiences mutations that detriment its fitness. Each of these operations are elaborated upon in the following subsections.

\subsubsection{Initialization}\label{subsubsec:initialization}

In order to seed the algorithm with initial candidate sets, we generate $P$ candidate sets $\bm{S^{'}}$ containing $K$ shapelets, with $K$ a random integer picked uniformly from $[2, W]$, $W$ a hyper-parameter of the algorithm, and $P$ the population size. $K$ is randomly chosen for each individual and the default value of $W$ is set to be $\sqrt{M}$. These two boundaries are chosen to be low in order to start with smaller candidate sets and grow them incrementally. This is beneficial for both the size of the final shapelet set as well as the runtime of each generation. For each candidate set we initialize, we randomly pick one of following two strategies with equal probability:
\begin{description}
	\item [\textbf{Initialization 1:}] apply $K$-means on a set of random subseries of a fixed random length sampled from~$\bm{T}$. The $K$ resulting centroids form a candidate set.
	\item [\textbf{Initialization 2:}] generate $K$ candidates of random lengths $(\in \{4,\ldots,max\_len\})$ by sampling them from~$\bm{T}$.
\end{description}

The $max\_len$ is a hyper-parameter that limits the length of the discovered shapelets, in order to combat overfitting. While \textbf{Initialization 1} results in strong initial individuals, \textbf{Initialization 2} is included in order to increase population diversity and to decrease the time required to initialize the entire population.

\subsubsection{Fitness}\label{subsubsec:fitness}

One of the most important components of a genetic algorithm, is its fitness function. In order to determine the fitness of a candidate set $\bm{S^{'}}$ we first construct $\bm{D^{'}}$, which is the distance matrix obtained by calculating the distances between $\bm{S^{'}}$ and $\bm{T}$. The goal of our genetic algorithm is find an $\bm{S^{'}}$ which produces a $\bm{D^{'}}$ that results in the most optimal predictive performance when provided to a classifier. We measure the predictive performance directly by means of an error function defined on the predictions of a logistic regression model and the provided label vector $\bm{y}$. When two candidate shapelet sets produce the same error, the set with the lowest complexity is deemed to be the fittest. The complexity of a shapelet set is expressed as the sum of shapelet lengths ($\sum_{s \in \bm{S}} |s|$). \\

The fitness calculation is the bottleneck of the algorithm. Calculating the distance of a shapelet with length $L$ to a time series of length $M$ requires $(M - L + 1) \times L$ pointwise comparisons. Thus, in the worst case, $\mathcal{O}(M^2)$ operations need to be performed per time series, resulting in a computational complexity of $\mathcal{O}(NM^2)$. We apply these distance calculations to each individual representing a collection of shapelets from our population, in each generation. Therefore, the complexity of the entire algorithm is equal to $\mathcal{O}(GPKNM^2)$, with $G$ the total number of generations, $P$ the population size, and $K$ the (maximum) number of shapelets in the bag each individual of the population represents.

\subsubsection{Crossover}\label{subsubsec:crossover}

\begin{figure*}
	\centering
	\includegraphics[width=1.0\textwidth]{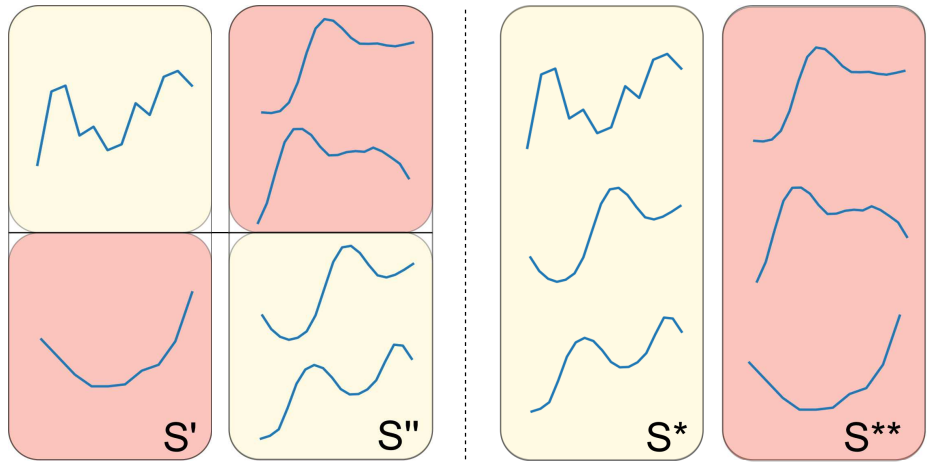}
	\caption{An example of a one-point crossover operation on two shapelet sets. Each original set is partitioned in two, and we take a partition from each set in order to construct a new set.}
	\label{fig:crossover1}      
\end{figure*}

\begin{figure*}
	\centering
	\includegraphics[width=1.0\textwidth]{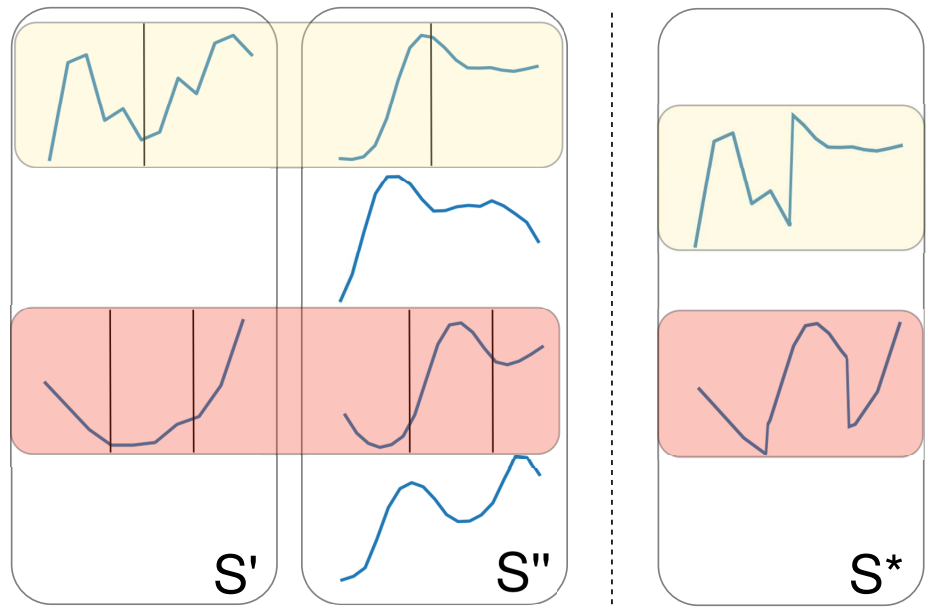}
	\caption{An example of one- and two-point crossover applied on individual shapelets.}
	\label{fig:crossover2}      
\end{figure*}

\begin{figure*}
	\centering
	\includegraphics[width=1.0\textwidth]{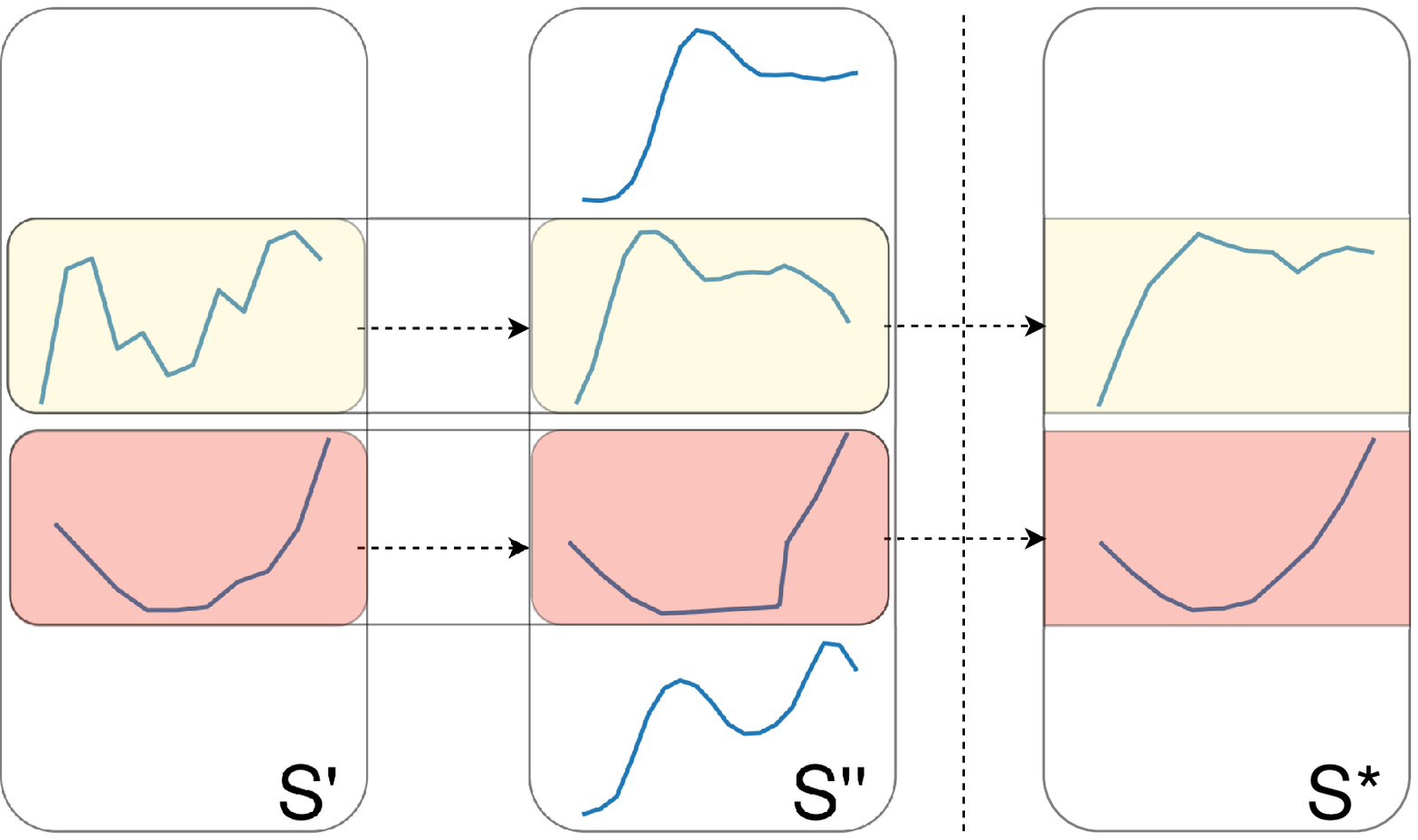}
	\caption{An example of the shapelet merging crossover operation.}
	\label{fig:crossover3}      
\end{figure*}

We define three different crossover operations, which take two candidate shapelet sets, $\bm{S^{'}}$ and $\bm{S^{''}}$, as input and produce two new sets, $\bm{S^{*}}$ and $\bm{S^{**}}$:
\begin{description}
	\item [\textbf{Crossover 1:}] apply one- or two-point crossover on two shapelet sets (each with a probability of 50\%). In other words, we create two new shapelet sets that are composed of shapelets from both $\bm{S^{'}}$ and $\bm{S^{''}}$. An example of this operation is provided in Figure~\ref{fig:crossover1}.
	\item [\textbf{Crossover 2:}] iterate over each shapelet $s$ in $\bm{S^{'}}$ and apply one- or two-point crossover (again with a probability of 50\%) with another randomly chosen shapelet from $\bm{S^{''}}$ to create $\bm{S^{*}}$. Apply the same, vice versa, to obtain $\bm{S^{**}}$. This differs from the first crossover operation as the one- or two-point crossover are performed on individual shapelets as opposed to entire sets. An example of this operation can be seen in Figure~\ref{fig:crossover2}.
	\item [\textbf{Crossover 3:}] iterate over each shapelet $s$ in $\bm{S^{'}}$ and merge it with another randomly chosen shapelet from $\bm{S^{''}}$. The merging of two shapelets can be done by calculating the mean (or barycenter) of the two time series. When two shapelets being merged have varying length, we merge the shorter shapelet with a random part of the longer shapelet.  A schematic overview of this strategy, on shapelets having the same length, is depicted in Figure~\ref{fig:crossover3}.
	
\end{description}
It is possible that all or no techniques are applied on a pair of individuals. Each technique has a probability equal to the configured crossover probability ($p_{\textrm{crossover}}$) of being applied.

\subsubsection{Mutations}\label{subsubsec:mutation}

The mutation operators are a vital part of the genetic algorithm, as they ensure population diversity and allow to escape from local optima in the search space. They take a candidate set $\bm{S^{'}}$ as input and produce a new, modified $\bm{S^{*}}$. In our approach, we define three simple mutation operators: 

\begin{description}
	\item [\textbf{Mutation 1:}] take a random $s \in \bm{S^{'}}$ and randomly remove a variable amount of data points from the beginning or ending of the time series.
	\item [\textbf{Mutation 2:}] remove a random $s \in \bm{S^{'}}$.
	\item [\textbf{Mutation 3:}] create a new candidate using \textbf{Initialization 2} and add it to $\bm{S^{'}}$.
\end{description}

Again, all techniques can be applied on a single individual, each having a probability equal to the configured mutation probability ($p_{\textrm{mutation}}$).

\subsubsection{Selection, Elitism \& Early Stopping}\label{subsubsec:selection}

After each generation, a fixed number of candidate sets are chosen based on their fitness for the next generation. Many different techniques exist to select these candidate sets. We chose to apply tournament selection with small tournament sizes. In this strategy, a number of candidate sets are sampled uniformly from the entire population to form a tournament. Afterwards, one candidate set is sampled from the tournament, where the probability of being sampled is determined by its fitness. Smaller tournament sizes ensure better population diversity as the probability of the fittest individual being included in the tournament decreases. Using this strategy, it is however possible that the fittest candidate set from the population is never chosen to compete in a tournament. Therefore, we apply elitism and guarantee that the fittest candidate set is always transferred to the next generation's population. Finally, since it can be hard to determine the ideal number of generations that a genetic algorithm should run, we implemented early stopping where the algorithm preemptively stops as soon as no candidate set with a better fitness has been found for a certain number of iterations ($patience$).

%
%
%
%
%
%
%
%
%
%
%
%
%
%

\subsubsection{List of all hyper-parameters}

We now present an overview of all hyper-parameters included in \textsc{\texttt{gendis}}, along with their corresponding explanation and default values.

\begin{itemize}
	\item Maximum shapelets per candidate ($W$): the maximum number of shapelets in a newly generated individual during initialization (default: $\sqrt{M}$).
	\item Population size ($P$): the total number of candidates that are evaluated and evolved in every iteration (default: $100$).
	\item Maximum number of generations ($G$): the maximum number of iterations the algorithm runs (default: $100$).
	\item Early stopping patience ($patience$): the algorithm preemptively stops evolving when no better individual has been found for $patience$ iterations (default: $10$).
	\item Mutation probability ($p_{\textrm{mutation}}$): the probability that a mutation operator gets applied to an individual in each iteration (default: $0.1$).
	\item Crossover probability ($p_{\textrm{crossover}}$): the probability that a crossover operator is applied on a pair of individuals in each iteration (default: $0.4$).
	\item Maximum shapelet length ($max\_len$): the maximum length of the shapelets in each shapelet set (individual). (default: $M$).
	\item The operations used during the initialization, crossover and mutation phases are configurable as well. (default: all mentioned operations).
\end{itemize}

\section{Results}\label{sec:results}

In the following subsections, we will present the setup of different experiments and the corresponding results in order to highlight the advantages of \textsc{\texttt{gendis}}.

\subsection{Efficiency of genetic operators}
In this section, we assess the efficiency of the introduced genetic operators by evaluating the fitness in function of the number of generations using different sets of operators. It should be noted that our implementation easily allows to configure the number and type of operators used for each of the different steps in the genetic algorithm, allowing the user to tune these according to the dataset.

\subsubsection{Datasets}
We pick six datasets, with varying characteristics, to evaluate the fitness of different configurations on. The chosen datasets, and their corresponding properties are summarized in Table~\ref{tab:datasets}.

\begin{table}[h!]
	\centering
	\scriptsize
	\begin{tabular}{lllll}
		\toprule
		Dataset  & \#Cls & TS\_len & \#Train & \#Test \\ \midrule
		ItalyPowerDemand          & 2         & 24        & 67      & 1029  \\
		SonyAIBORobotSurface2     & 2         & 65        & 27      & 953   \\
		FaceAll                   & 14        & 131       & 560     & 1690  \\
		Wine                      & 2         & 234       & 57      & 54  \\
		PhalangesOutlinesCorrect  & 2         & 80        & 1800    & 858  \\
		Herring                   & 2         & 512       & 64      & 64  \\ \bottomrule
	\end{tabular}
	\caption{The chosen datasets, having varying characteristics, for the evaluation of the genetic operators' efficiency. \#Cls = number of classes, TS\_len = length of time series, \#Train = number of training time series, \#Train = number of testing time series}
	\label{tab:datasets}
\end{table}

\subsubsection{Initialization operators}
\begin{figure*}
	\centering
	\includegraphics[width=1.0\textwidth]{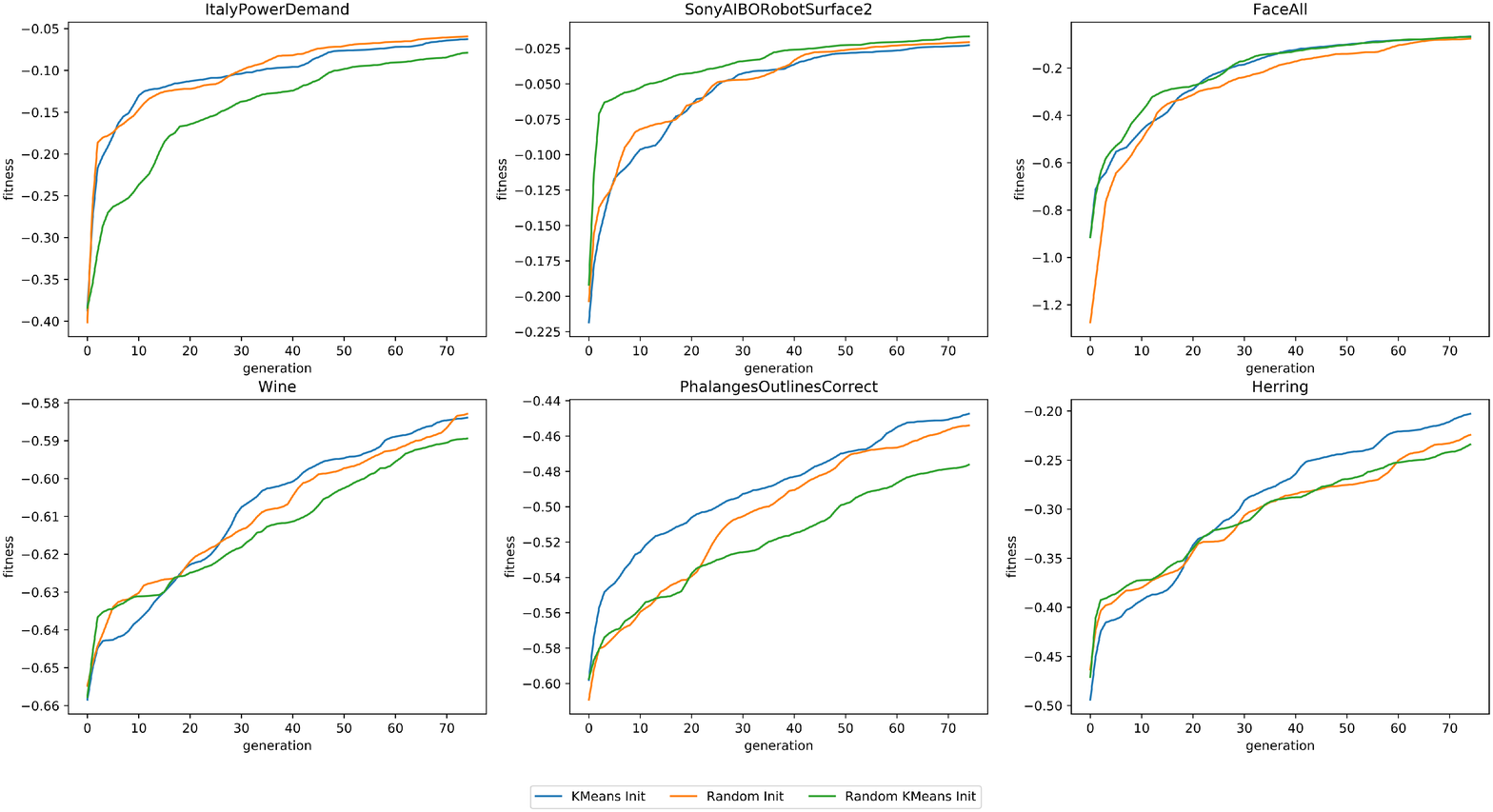}
	\caption{The fitness in function of the number of generations, for six datasets, using three different configurations of initialization operations.}
	\label{fig:initializations}      
\end{figure*}

We first compare the fitness of GENDIS using three different sets of initialization operators:
\begin{itemize}
	\item Initializing the individuals with K-Means (Initialization 1)
	\item Randomly initializing the shapelet sets (Initialization 2)
	\item Using all two initialization operations
\end{itemize}
Each configuration was tested using a small population (25 individuals), in order to reduce the required computational time, for 75 generations, as the impact of the initialization is highest in the earlier generations. All mutation and crossover operators were used. We show the average fitness of all individuals in the population in Figure~\ref{fig:initializations}. From these results, we can conclude that the two initialization operators are competitive to each other, as one operator will outperform the other on several datasets and vice versa on the others.

\subsubsection{Crossover operators}
We now compare the average fitness of all individuals in the population, in function of the number of generations, when configuring GENDIS to use four different sets of crossover operators:
\begin{itemize}
	\item Using solely point crossovers on the shapelet sets (Crossover 1)
	\item Using solely point crossovers on individual shapelets (Crossover 2)
	\item Using solely merge crossovers (Crossover 3)
	\item Using all three crossover operations
\end{itemize}

\begin{figure*}
	\centering
	\includegraphics[width=1.0\textwidth]{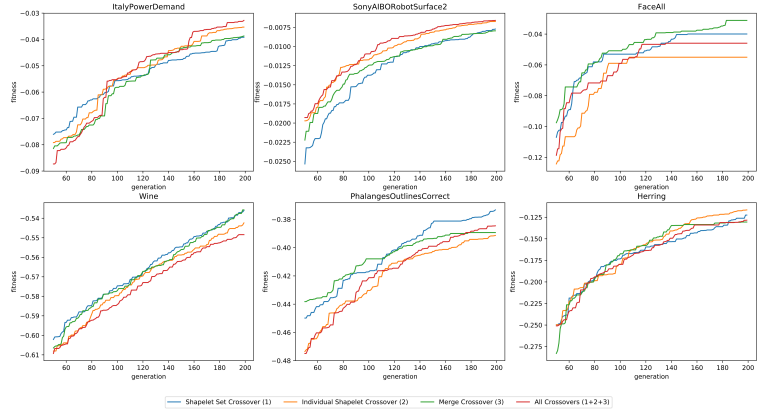}
	\caption{The fitness in function of the number of generations, for six datasets, using four different configurations of crossover operations.}
	\label{fig:crossovers}      
\end{figure*}

Each run had a population of 25 individuals and ran for 200 generations. All mutation and initialization operators were used. As the average fitness is rather similar in the earlier generations, we truncate the first 50 measurements to better highlight the differences. The results are presented in Figure~\ref{fig:crossovers}. As can be seen, it is again difficult to single out an operation that significantly outperforms the others.

\subsubsection{Mutation operators}
\begin{figure*}
	\centering
	\includegraphics[width=1.0\textwidth]{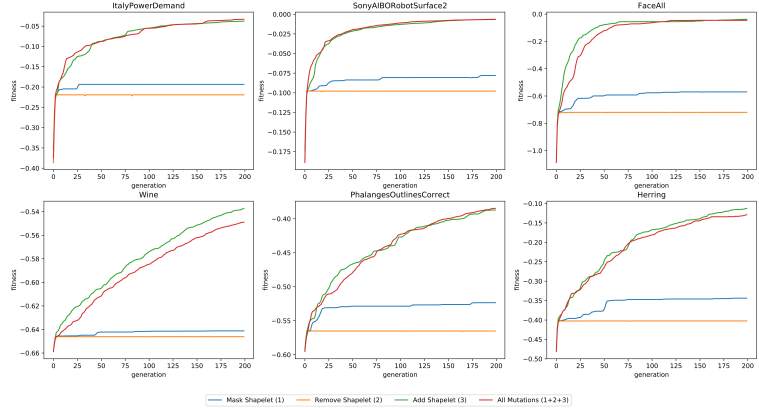}
	\caption{The fitness in function of the number of generations, for six datasets, using four different configurations of mutation operations.}
	\label{fig:mutations}      
\end{figure*}

The same experiment was performed to assess the efficiency of the mutation operators. Four different configurations were used:
\begin{itemize}
	\item Masking a random part of a shapelet (Mutation 1)
	\item Removing a random shapelet from the set (Mutation 2)
	\item Adding a shapelet, randomly sampled from the data, to the set (Mutation 3)
	\item Using all three mutation operations
\end{itemize}
The average fitness of the individuals, in function of the number of generations is depicted in Figure~\ref{fig:mutations}. It is clear that the addition of shapelets (Mutation 3) is the most significant operator. Without it, the fitness quickly converges to a sub-optimal value. The removal and masking of shapelets does not seem to increase the average fitness often, but are important operators in order to keep the the number of shapelets and the length of the shapelets small.

\subsection{Evaluating sets of candidates versus single candidates} \label{subsec:sets}

\begin{figure*}[h!]
	\centering
	\includegraphics[width=0.95\textwidth]{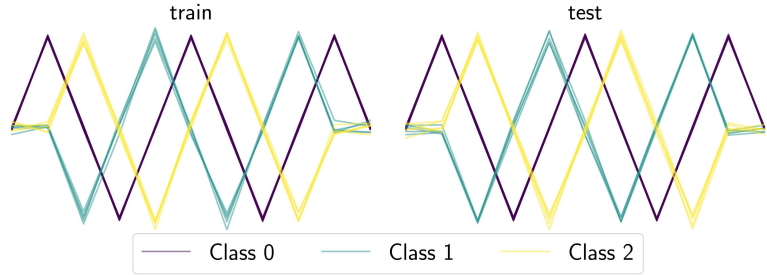}
	\caption{The generated train and test set for the artificial classification problem.}
	\label{fig:artificial_data}      
\end{figure*}

A key factor of \textsc{\texttt{gendis}}, is that it evaluates entire sets of shapelets (a dependency between the shapelets is introduced), as opposed to evaluating single candidates independently and taking a top-k. The disadvantage of the latter approach is that similar shapelets will achieve similar values given a certain metric. When entire sets are evaluated, we can both optimize a quality metric for candidate sets, as the size of each of these sets. This results in smaller sets with less similar shapelets. Moreover, interactions between shapelets can be explicitly taken into account. To demonstrate these advantages, we compare \textsc{\texttt{gendis}} to \textsc{\texttt{st}}, which evaluates candidate shapelets individually, on an artificial three-class dataset, depicted in Figure~\ref{fig:artificial_data}. The constructed dataset contains a large number of very similar time series of class 0, while having a smaller number of more dissimilar time series of class 1 and 2. The distribution of time series across the three classes in both the train and test dataset is thus skewed, with the number of samples in class $0, 1, 2$ being equal to $25, 5, 5$ respectively. This imbalance causes the independent approach to focus solely on extracting shapelets that can discriminate class 0 from the two others, since the information gain will be highest for these individual shapelets. Clearly, this is not ideal as subsequences taken from time series of class 0 possess little to no discriminative power for the two other classes, as the distances to time series from these two classes will be nearly equal. \\

\begin{figure*}[h!]
	\centering
	\includegraphics[width=0.95\textwidth]{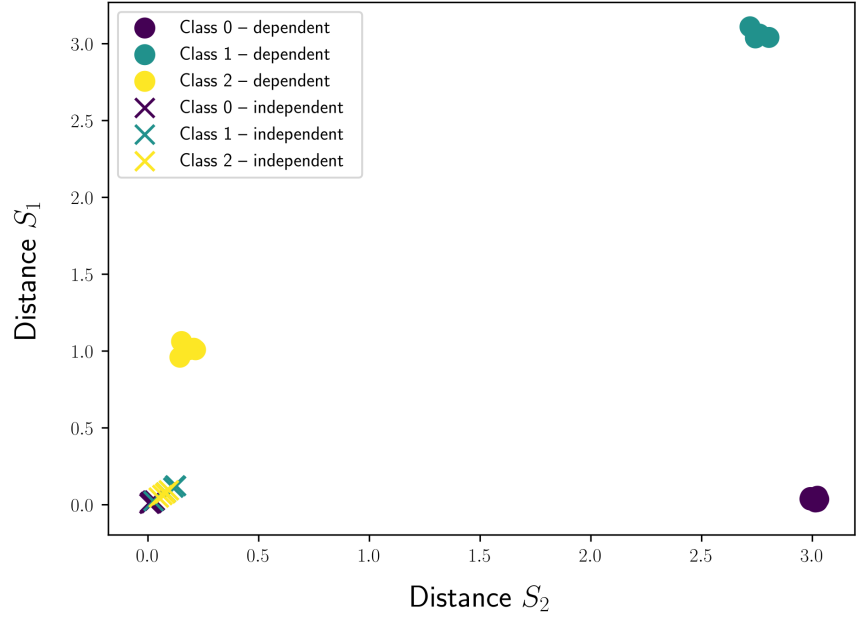}
	\caption{The samples of the test set are represented by markers (circles for \textsc{\texttt{gendis}}) and crosses for \textsc{\texttt{st}}) while the axes correspond to the distance to a shapelet.}
	\label{fig:distances_scatter}      
\end{figure*}

We extract two shapelets with both techniques, which allows us to visualize the different test samples in a two-dimensional transformed distance space, as shown in Figure~\ref{fig:distances_scatter}. Each axis of this space represents the distances to a certain shapelet. For the independent approach, we can clearly see that the distances of the samples for all three classes to both shapelets are clustered near the origin of the space, making it very hard for a classifier to draw a separation boundary. On the other hand, a very clear separation can be seen for the samples of the three classes when using the shapelets discovered by \textsc{\texttt{gendis}}, a dependent approach. The low discriminative power of the indepent approach is confirmed by fitting a Logistic Regression model with tuned regularization type and strength on the obtained distances. The classifier fitted on the distances extracted by the independent approach is only able to achieve an accuracy of $0.8286$ ($\frac{29}{35}$) on the rather straight-forward dataset. The accuracy score of \textsc{\texttt{gendis}}, a dependent approach, equals $1.0$.

\subsection{Discovering shapelets outside the data} \label{subsec:outside}
\begin{figure*}
	\centering
	\begin{subfigure}{.45\textwidth}
		\centering
		\includegraphics[width=1\linewidth]{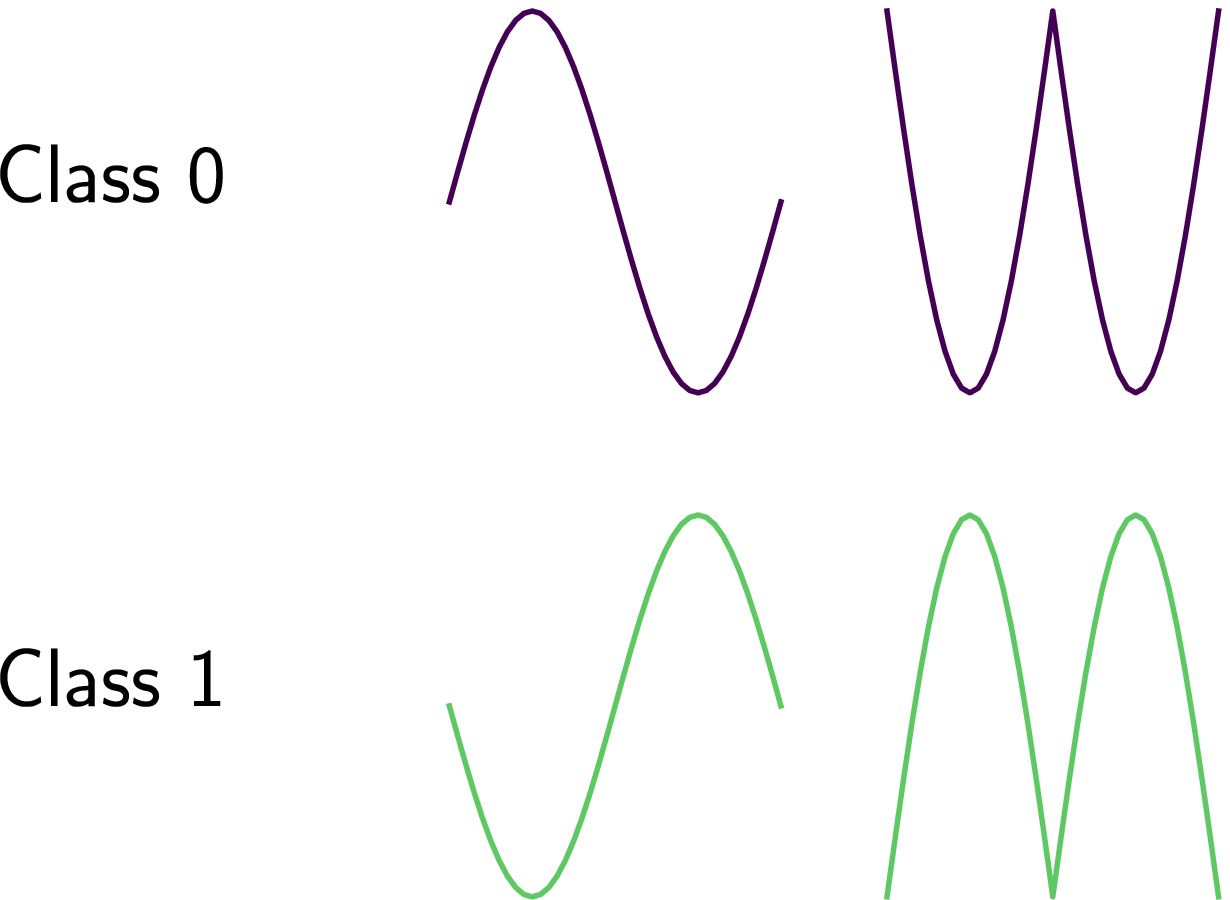}
		\caption{}
		\label{fig:unlimited_discovery_sub1}
	\end{subfigure}
	\hfill
	\begin{subfigure}{.45\textwidth}
		\centering
		\includegraphics[width=1\linewidth]{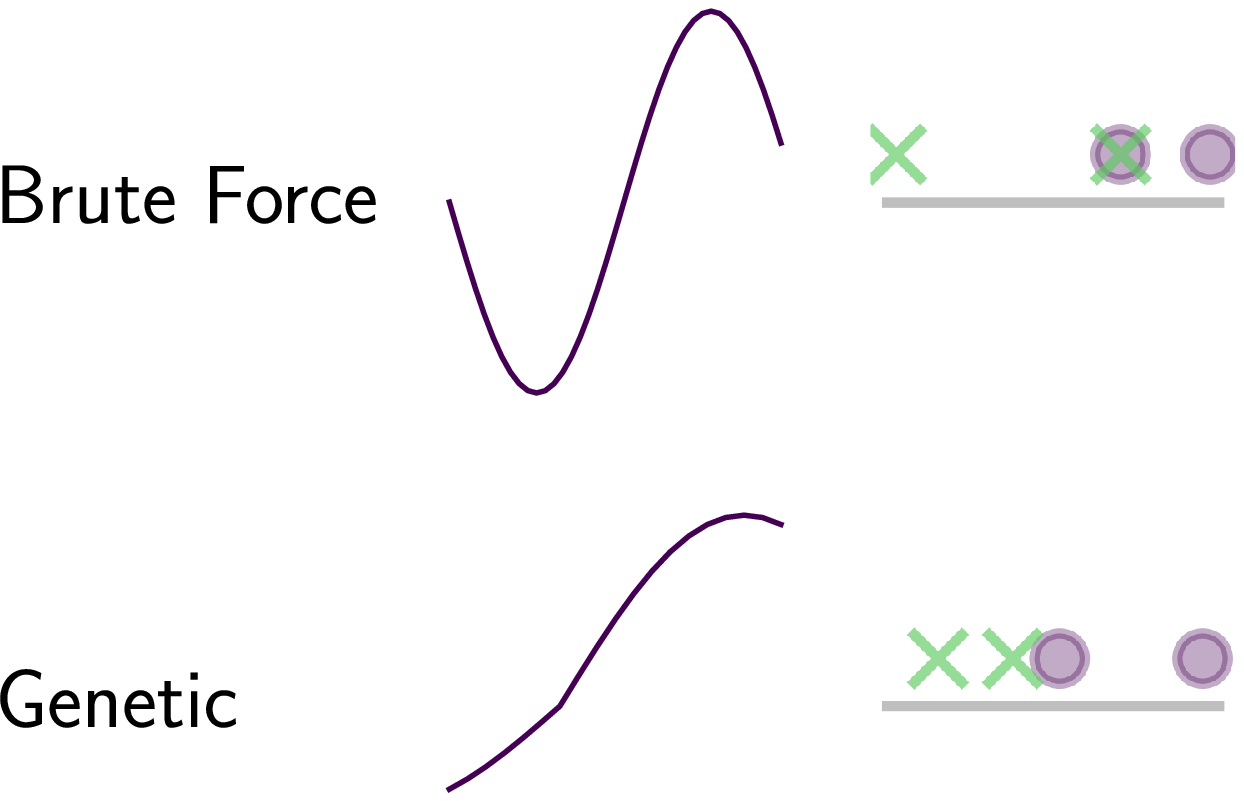}
		\caption{}
		\label{fig:unlimited_discovery_sub2}
	\end{subfigure}
	\caption{A two-class problem with two time series per class and the extracted shapelets with corresponding distances on an ordered line by a brute-force approach versus \textsc{\texttt{gendis}}. The crosses on the ordered line correspond to distances of the shapelet to the time series from Class 1 while the circles on the ordered line correspond to distances to Class 0 (the more to the right, the higher the distance).}
	\label{fig:unlimited_discovery}
\end{figure*}

Another advantage of \textsc{\texttt{gendis}}, is that the discovery of shapelets is not limited to be a subseries from $\bm{T}$. Due to the nature of the evolutionary process, the discovered shapelets can have a distance greater than 0 to all time series in the dataset. More formally: $\exists s \in \bm{S}.\ \forall t \in \bm{T}.\ dist(s, t) > 0$. While this can be somewhat detrimental concerning interpretability, it can be necessary to get an excellent predictive performance. We demonstrate this through a very simple, artificial example. Assume we have a two-class classification problem and are provided two time series per class, as illustrated in Figure~\ref{fig:unlimited_discovery_sub1}. The extracted shapelet, and the distances to each time series, by a brute force approach and a slightly modified version of \textsc{\texttt{gendis}} can be found in Figure~\ref{fig:unlimited_discovery_sub2}. The modification we made to \textsc{\texttt{gendis}} is that we specifically search for only one shapelet instead of an entire set of shapelets. We can see that the exhaustive search approach is not able to find a subseries in any of these four time series that separates both classes while the shapelet extracted by \textsc{\texttt{gendis}} ensures perfect separation. \\

It is important to note here that discovering shapelets outside the data sacrifices interpretability for an increase in predictive performance of the shapelets. As the operators that are used during the genetic algorithm are completely configurable for \textsc{\texttt{gendis}}, one can use only the first crossover operation (one- or two-point crossover on shapelet sets) to ensure all shapelets come from within the data.

\subsection{Stability}
\begin{figure*}[h!]
	\centering
	\includegraphics[width=0.8\textwidth]{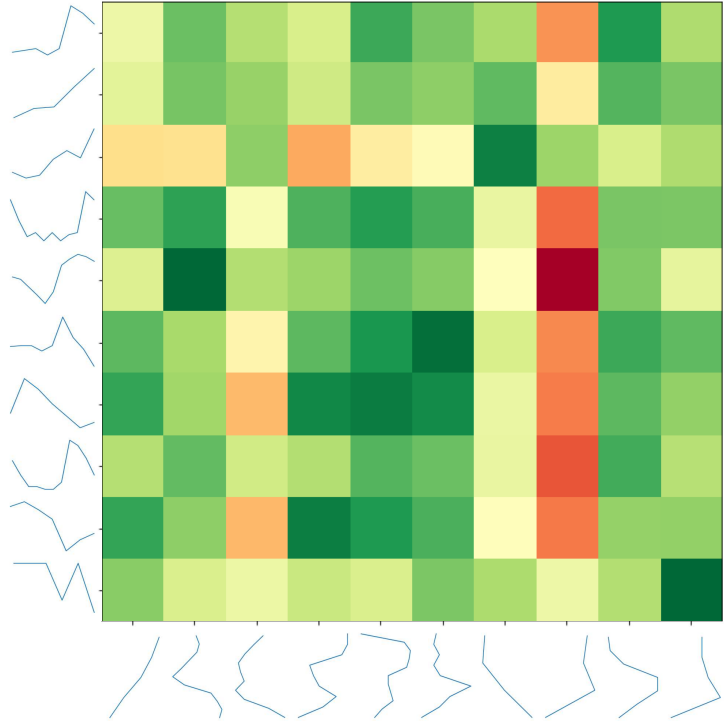}
	\caption{A pairwise distance matrix, constructed using Dynamic Time Warping, between discovered shapelet sets of two different runs on the ItalyPowerDemand dataset.}
	\label{fig:stability}      
\end{figure*}

In order to evaluate the stability of our algorithm, we compare the extracted shapelets of two different runs on the ItalyPowerDemand dataset. We set the algorithm to evolve a large population (100 individuals) for a large number of generations (500) in order to ensure convergence. Moreover, we limit the maximum number of extracted shapelets to 10, in order to keep the visualization clear. We then calculated the similarity of the discovered shapelets between the two runs, using Dynamic Time Warping~\citep{berndt1994using}. A heatmap of the distances is depicted in Figure~\ref{fig:stability}. While the discovered shapelets are not exactly the same, we can often find pairs that contain the same semantic intelligence, such as a saw pattern or straight lines.

\subsection{Comparing \textsc{\texttt{gendis}} to \textsc{\texttt{fs}}, \textsc{\texttt{st}} and \textsc{\texttt{lts}}} \label{subsec:comparison}

In this section, we compare our algorithm \textsc{\texttt{gendis}} to the results from~\citep{bagnall2016great}, which are hosted online\footnote{ \url{www.time seriesclassification.com}}. In that study, 31 different algorithms, including three shapelet discovery techniques, have been compared on 85 datasets. The 85 datasets stem from different data sources and different domains, including electrocardiogram data from the medical domain and sensor data from the IoT domain. The three included shapelet techniques are Shapelet Transform (\textsc{\texttt{st}})~\citep{lines2012shapelet}, Learning Time Series Shapelets (\textsc{\texttt{lts}})~\citep{grabocka2014learning}, and Fast Shapelets (\textsc{\texttt{fs}})~\citep{rakthanmanon2013fast}. A discussion of all three techniques can be found in Section~\ref{sec:related_work}. \\

\begin{figure*}[h!]
	\centering
	\includegraphics[width=1.0\textwidth]{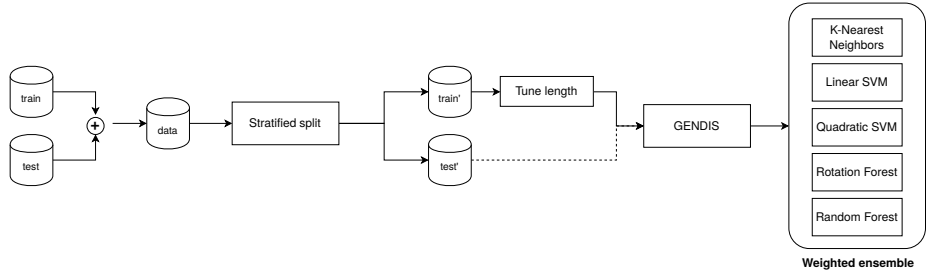}
	\caption{The evaluation setup used to compare \textsc{\texttt{gendis}} to other shapelet techniques. Train and test data are first concatenated and then re-distributed to form new train and test sets with similar distributions to the ones prior to the concatenation. The newly created train set is then used to tune the optimal maximal length of the shapelets in cross-validation and used to extract shapelets and to train the ensemble of classifiers. Finally, the test set is used to evaluate the extracted shapelets and fitted ensemble.}
	\label{fig:evaluation_setup}      
\end{figure*}

For 84 of the 85 datasets, we conducted twelve measurements by concatenating the provided training and testing data and re-partitioning in a stratified manner, as done in the original study. Only the `Phoneme' dataset could not be included due to problems downloading the data while executing this experiment. On every dataset, we used the same hyper-parameter configuration for \textsc{\texttt{gendis}}: a population size of 100, a maximum of 100 iterations, early stopping after 10 iterations and crossover \& mutation probabilities of $0.4$ and $0.1$ respectively. The only parameter that was tuned for every dataset separately was a maximum length for each shapelet, to combat overfitting. To tune this, we picked the length $l \in [\frac{M}{4}, \frac{M}{2}, \frac{3M}{4}, M]$ that resulted in the best logarithmic (or entropy) loss using 3-fold cross validation on the training set. The distance matrix obtained through the extracted shapelets of \textsc{\texttt{gendis}} was then fed to a heterogeneous ensemble consisting of a rotation forest, random forest, support vector machine with linear kernel, support vector with quadratic kernel and a k-nearest neighbor classifier~(\cite{large2017heterogeneous}). This ensemble matches the one used by the best-performing algorithm, \textsc{\texttt{st}}, closely. This is in contrast with \textsc{\texttt{fs}} which produces a decision tree and \textsc{\texttt{lts}} which learns a separating hyperplane (similar to logistic regression) jointly with the shapelets. This setup is also depicted schematically in Figure~\ref{fig:evaluation_setup}. Trivially, the ensemble will empirically outperform each of the individual classifiers~\citep{dietterich2000ensemble}, but it does take a longer time to fit and somewhat takes the focus away from the quality of the extracted shapelets. Nevertheless, it is necessary to use an ensemble in order to allow for a fair comparison with \textsc{\texttt{st}}, as that was used by~\cite{bagnall2016great} to generate their results. To give more insights into the quality of the extracted shapelets, we also report the accuracies using a Logistic Regression classifier. We tuned the type of regularization (Ridge vs Lasso) and the regularization strength ($C \in \{0.001, 0.01, 0.1, 1.0, 10.0, 100.0, 1000.0\}$) using the training set. We recommend future research to compare their results to those obtained with Logistic Regression classifier.  \\

The mean accuracy over the twelve measurements of \textsc{\texttt{gendis}} in comparison to the mean of the hundred original measurements of the three other algorithms, retrieved from the online repository, can be found in Table~\ref{tbl:gendis_all_1}~and~\ref{tbl:gendis_all_2}. While a smaller number of measurements is conducted within this study, it should be noted that the measurements from~\cite{bagnall2016great} took over 6 months to generate. Moreover, accuracy is often not the most ideal metric to measure the predictive performance with. Although it is one of the most intuitive metrics, it has several disadvantages such as skewness when data is imbalanced. Nevertheless, the accuracy metric is the only one allowing for comparison to related work, as that metric was used in those studies. Moreover, the used datasets are merely benchmark datasets and the goal is solely to compare the quality of the shapelets extracted by \textsc{\texttt{gendis}} to those of \textsc{\texttt{st}}. We recommend to use different performance metrics, which should be tailored to the specific use case. An example is using the area under the receiver operating characteristic curve (AUC) in combination with precision and recall for medical datasets. \\

For each dataset, we also perform an unpaired student t-test with a cutoff value of $0.05$ to detect statistically significant differences. When the performance of an algorithm for a certain dataset is statistically better than all others, it is indicated in bold. From these results, we can conclude that \textsc{\texttt{fs}} is inferior to the three other techniques while \textsc{\texttt{st}} most often achieves the best performance, but at a very high computational complexity. \\

\begin{figure*}[h!]
	\centering
	\includegraphics[width=1.0\textwidth]{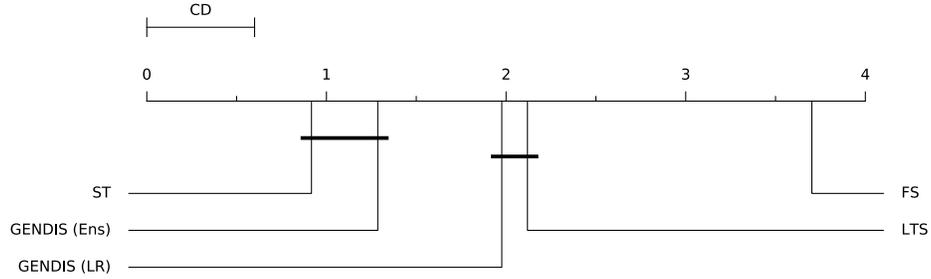}
	\caption{A critical difference diagram of the four evaluated shapelet discovery algorithms. For each technique, the average rank is calculated. Then cliques (bold black lines) are formed if the $p$-value of the Holm post-hoc test is lower than $0.1$.}
	\label{fig:critical_difference}      
\end{figure*}

The average number of shapelets extracted by \textsc{\texttt{gendis}} is reported in the final column. The number of shapelets extracted by \textsc{\texttt{st}} in the original study equals $10*N$. Thus, the total number of shapelets used to transform the original time series to distances is at least an order of magnitude less when using \textsc{\texttt{gendis}}. In order to compare the algorithms across all datasets, a Friedman ranking test~(\cite{friedman1937use}), was applied with a Holm post-hoc correction~(\cite{holm1979simple,benavoli2016should}). We present the average rank of each algorithm using a critical difference diagram, with cliques formed using the results of the Friedman test with a Holm post-hoc correction at a significance cutoff level of $0.1$ in Figure~\ref{fig:critical_difference}. The higher the cutoff level, the less probable it is to form cliques. For \textsc{\texttt{gendis}}, both the results obtained with the ensemble and with the logistic regression classifier are used. From this, we can conclude that there is no statistical difference between \textsc{\texttt{st}} and \textsc{\texttt{gendis}} while both are statistically better than \textsc{\texttt{fs}} and \textsc{\texttt{lts}}. \\

\begin{table}[t]
	\centering
	\scriptsize
	\begin{tabular}{>{\tiny}lllll|lllll|l}
		\toprule
		\multirow{2}{*}{\scriptsize Dataset} & \multirow{2}{*}{\#Cls} & \multirow{2}{*}{TS\_len} & \multirow{2}{*}{\#Train} & \multirow{2}{*}{\#Test} & \multicolumn{2}{c}{GENDIS} & \multirow{2}{*}{ST} & \multirow{2}{*}{LTS} & \multirow{2}{*}{FS} & \multirow{2}{*}{\#Shaps} \\
		& & & & & Ens & LR & & & & \\ \midrule
		Adiac                        & 37        & 176       & 390     & 391    & 66.2  & 69.8  & \textbf{76.8} & 42.9          & 55.5 & 39          \\
		ArrowHead                    & 3         & 251       & 36      & 175    & 79.4  & 82.0  & 85.1          & 84.1          & 67.5 & 39          \\
		Beef                         & 5         & 470       & 30      & 30     & 51.5  & 58.8  & \textbf{73.6} & 69.8          & 50.2 & 41          \\
		BeetleFly                    & 2         & 512       & 20      & 20     & 90.6  & 87.5  & 87.5          & 86.2          & 79.6 & 42          \\
		BirdChicken                  & 2         & 512       & 20      & 20     & 90.0  & 90.5  & 92.7          & 86.4          & 86.2 & 45          \\
		CBF                          & 3         & 128       & 30      & 900    & 99.1  & 97.6  & 98.6          & 97.7          & 92.4 & 43          \\
		Car                          & 4         & 577       & 60      & 60     & 82.5  & 83.0  & \textbf{90.2} & 85.6          & 73.6 & 48          \\
		ChlorineConcentration        & 3         & 166       & 467     & 3840   & 60.9  & 57.5  & \textbf{68.2} & 58.6          & 56.6 & 30          \\
		CinCECGTorso                 & 4         & 1639      & 40      & 1380   & 92.1  & 91.5  & 91.8          & 85.5          & 74.1 & 57          \\
		Coffee                       & 2         & 286       & 28      & 28     & 98.9  & 98.6  & 99.5          & 99.5          & 91.7 & 44          \\
		Computers                    & 2         & 720       & 250     & 250    & 75.4  & 72.7  & \textbf{78.5} & 65.4          & 50.0 & 38          \\
		CricketX                     & 12        & 300       & 390     & 390    & 73.1  & 66.6  & \textbf{77.7} & 74.4          & 47.9 & 41          \\
		CricketY                     & 12        & 300       & 390     & 390    & 69.8  & 64.3  & \textbf{76.2} & 72.6          & 50.9 & 41          \\
		CricketZ                     & 12        & 300       & 390     & 390    & 72.6  & 65.9  & \textbf{79.8} & 75.4          & 46.6 & 40          \\
		DiatomSizeReduction          & 4         & 345       & 16      & 306    & \textbf{97.4} & 96.5 & 91.1          & 92.7          & 87.3 & 45          \\
		DistalPhalanxOutlineAgeGroup & 3         & 80        & 400     & 139    & \textbf{84.4} & 83.2 & 81.9          & 81.0          & 74.5 & 32          \\
		DistalPhalanxOutlineCorrect  & 2         & 80        & 600     & 276    & 82.4  & 81.5        & 82.9          & 82.2          & 78.0 & 32          \\
		DistalPhalanxTW              & 6         & 80        & 400     & 139    & \textbf{76.7} & 76.0 & 69.0          & 65.9          & 62.3 & 33          \\
		ECG200                       & 2         & 96        & 100     & 100    & 86.3 & 86.5         & 84.0          & 87.1          & 80.6 & 30          \\
		ECG5000                      & 5         & 140       & 500     & 4500   & 94.0  & 93.8        & \textbf{94.3} & 94.0          & 92.2 & 34          \\
		ECGFiveDays                  & 2         & 136       & 23      & 861    & 99.9 & \textbf{100.0} & 95.5          & 98.5          & 98.6 & 33          \\
		Earthquakes                  & 2         & 512       & 322     & 139    & \textbf{78.4} & 73.7 & 73.7          & 74.2          & 74.7 & 44          \\
		ElectricDevices              & 7         & 96        & 8926    & 7711   & 83.7  & 77.6         & \textbf{89.5} & 70.9          & 26.2 & 31          \\
		FaceAll                      & 14        & 131       & 560     & 1690   & 94.5 & 92.6          & \textbf{96.8} & 92.6          & 77.2 & 38          \\
		FaceFour                     & 4         & 350       & 24      & 88     & 93.2 & 94.1          & 79.4          & \textbf{95.7} & 86.9 & 41          \\
		FacesUCR                     & 14        & 131       & 200     & 2050   & 90.1 & 89.0         & 90.9          & \textbf{93.9} & 70.1 & 39          \\
		FiftyWords                   & 50        & 270       & 450     & 455    & \textbf{72.9} & 71.8 & 71.3          & 69.4          & 51.2 & 39          \\
		Fish                         & 7         & 463       & 175     & 175    & 87.0 & 90.5         & \textbf{97.4} & 94.0          & 74.2 & 50          \\
		FordA                        & 2         & 500       & 3601    & 1320   & 90.8 & 90.7         & \textbf{96.5} & 89.5          & 78.5 & 37          \\
		FordB                        & 2         & 500       & 3636    & 810    & 89.5 & 89.8          & \textbf{91.5} & 89.0          & 78.3 & 38          \\
		GunPoint                     & 2         & 150       & 50      & 150    & 96.9 & 95.7         & \textbf{99.9} & 98.3          & 93.0 & 39          \\
		Ham                          & 2         & 431       & 109     & 105    & 72.9 & 77.2          & 80.8          & \textbf{83.2} & 67.7 & 37          \\
		HandOutlines                 & 2         & 2709      & 1000    & 370    & 89.7 & 91.0         & \textbf{92.4} & 83.7          & 84.1 & 41          \\
		Haptics                      & 5         & 1092      & 155     & 308    & 45.2 & 43.9         & \textbf{51.2} & 47.8          & 35.6 & 55          \\
		Herring                      & 2         & 512       & 64      & 64     & 59.6 & 61.8         & \textbf{65.3} & 62.8          & 55.8 & 42          \\
		InlineSkate                  & 7         & 1882      & 100     & 550    & \textbf{43.8} & 39.3 & 39.3          & 29.9          & 25.7 & 58          \\
		InsectWingbeatSound          & 11        & 256       & 220     & 1980   & 57.3 & 57.5         & \textbf{61.7} & 55.0          & 48.8 & 36          \\
		ItalyPowerDemand             & 2         & 24        & 67      & 1029   & 95.6 & \textbf{96.0}         & 95.3          & 95.2          & 90.9 & 31          \\
		LargeKitchenAppliances       & 3         & 720       & 375     & 375    & 91.0 & 90.4          & \textbf{93.3} & 76.5          & 41.9 & 33          \\
		Lightning2                   & 2         & 637       & 60      & 61     & \textbf{80.9} & 79.1 & 65.9          & 75.9          & 48.0 & 39          \\
		Lightning7                   & 7         & 319       & 70      & 73     & \textbf{78.2} & 76.3          & 72.4          & 76.5          & 10.1 & 39          \\
		Mallat                       & 8         & 1024      & 55      & 2345   & \textbf{98.2} & 97.3 & 97.2          & 95.1          & 89.3 & 58          \\
		
		\bottomrule
	\end{tabular}
	\caption{A comparison between \textsc{\texttt{gendis}} and three other shapelet techniques on 85 datasets. For each dataset, we report the total number of classes, the length of the time series, the number of time series in the train and test set, the accuracy score on the test set achieved by \textsc{\texttt{gendis}} (using both an ensemble (Ens) and logistic regression (LR)), \textsc{\texttt{st}}, \textsc{\texttt{lts}} \& \textsc{\texttt{fs}}, and finally the average number of shapelets extracted by \textsc{\texttt{gendis}}. When a technique is statistically significant better than the others, according to a student t-test with a cutoff of $0.05$, it is marked as bold.}
	\label{tbl:gendis_all_1}
\end{table}

\begin{table}[t]
	\centering
	\scriptsize
	\begin{tabular}{>{\tiny}lllll|lllll|l}
		\toprule
		\multirow{2}{*}{\scriptsize Dataset} & \multirow{2}{*}{\#Cls} & \multirow{2}{*}{TS\_len} & \multirow{2}{*}{\#Train} & \multirow{2}{*}{\#Test} & \multicolumn{2}{c}{GENDIS} & \multirow{2}{*}{ST} & \multirow{2}{*}{LTS} & \multirow{2}{*}{FS} & \multirow{2}{*}{\#Shaps} \\
		& & & & & Ens & LR & & & & \\ \midrule
		Meat                           & 3         & 448       & 60      & 60     & 98.7 & \textbf{98.8}  & 96.6           & 81.4          & 92.4  & 48          \\
		MedicalImages                  & 10        & 99        & 381     & 760    & \textbf{72.4} & 68.6  & 69.1           & 70.4          & 60.9  & 37          \\
		MiddlePhalanxOutlineAgeGroup   & 3         & 80        & 400     & 154    & \textbf{74.4} & 73.2  & 69.4           & 67.9          & 61.3  & 30          \\
		MiddlePhalanxOutlineCorrect    & 2         & 80        & 600     & 291    & 80.7 & 79.6          & 81.5           & \textbf{82.2} & 71.6  & 30          \\
		MiddlePhalanxTW                & 6         & 80        & 399     & 154    & 62.2 & \textbf{63.1}  & 57.9           & 54.0          & 51.9  & 37          \\
		MoteStrain                     & 2         & 84        & 20      & 1252   & 86.3 & 86.6          & 88.2           & 87.6          & 79.3  & 36          \\
		NonInvasiveFatalECGThorax1     & 42        & 750       & 1800    & 1965   & 84.7 & 89.4          & \textbf{94.7}  & 60.0          & 71.0  & 41          \\
		NonInvasiveFatalECGThorax2     & 42        & 750       & 1800    & 1965   & 87.1 & 92.3          & \textbf{95.4}  & 73.9          & 75.8  & 37          \\
		OSULeaf                        & 6         & 427       & 200     & 242    & 76.2 & 75.8           & \textbf{93.4}  & 77.1          & 67.9  & 45          \\
		OliveOil                       & 4         & 570       & 30      & 30     & 86.1 & 88.8          & 88.1           & 17.2          & 76.5  & 53          \\
		PhalangesOutlinesCorrect       & 2         & 80        & 1800    & 858    & \textbf{80.8} & 78.7  & 79.4           & 78.3          & 73.0  & 30          \\
		Plane                          & 7         & 144       & 105     & 105    & 99.2 & 99.3           & \textbf{100.0} & 99.5          & 97.0  & 34          \\
		ProximalPhalanxOutlineAgeGroup & 3         & 80        & 400     & 205    & 84.1 & 83.9          & 84.1           & 83.2          & 79.7  & 32          \\
		ProximalPhalanxOutlineCorrect  & 2         & 80        & 600     & 291    & 86.2 & 85.7          & \textbf{88.1}  & 79.3          & 79.7  & 30          \\
		ProximalPhalanxTW              & 6         & 80        & 400     & 205    & \textbf{81.7} & 80.2  & 80.3           & 79.4          & 71.6  & 31          \\
		RefrigerationDevices           & 3         & 720       & 375     & 375    & 68.6 & 62.4           & \textbf{76.1}  & 64.2          & 57.4  & 34          \\
		ScreenType                     & 3         & 720       & 375     & 375    & 52.8 & 52.8          & \textbf{67.6}  & 44.5          & 36.5  & 37          \\
		ShapeletSim                    & 2         & 500       & 20      & 180    & \textbf{100.0} & 100.0 & 93.4           & 93.3          & 100.0 & 34          \\
		ShapesAll                      & 60        & 512       & 600     & 600    & 79.6 & 79.3           & \textbf{85.4}  & 76.0          & 59.8  & 44          \\
		SmallKitchenAppliances         & 3         & 720       & 375     & 375    & 74.3 & 74.2          & \textbf{80.2}  & 66.3          & 33.3  & 37          \\
		SonyAIBORobotSurface1          & 2         & 70        & 20      & 601    & \textbf{96.0} & 95.6  & 88.8           & 90.6          & 91.8  & 34          \\
		SonyAIBORobotSurface2          & 2         & 65        & 27      & 953    & 88.6 & 87.0           & \textbf{92.4}  & 90.0          & 84.9  & 34          \\
		StarLightCurves                & 3         & 1024      & 1000    & 8236   & 95.9 & 95.3          & \textbf{97.7}  & 88.8          & 90.8  & 30          \\
		Strawberry                     & 2         & 235       & 613     & 370    & 95.2 & 95.1           & \textbf{96.8}  & 92.5          & 91.7  & 34          \\
		SwedishLeaf                    & 15        & 128       & 500     & 625    & 88.7 & 87.7          & \textbf{93.9}  & 89.9          & 75.8  & 37          \\
		Symbols                        & 6         & 398       & 25      & 995    & \textbf{93.4} & 92.5  & 86.2           & 91.9          & 90.8  & 43          \\
		SyntheticControl               & 6         & 60        & 300     & 300    & 98.8 & 98.7           & 98.7           & \textbf{99.5} & 92.0  & 39          \\
		ToeSegmentation1               & 2         & 277       & 40      & 228    & 92.0 & 90.7          & \textbf{95.4}  & 93.4          & 90.4  & 40          \\
		ToeSegmentation2               & 2         & 343       & 36      & 130    & 93.1 & 90.6           & 94.7           & 94.3          & 87.3  & 36          \\
		Trace                          & 4         & 275       & 100     & 100    & 100.0 & 99.9          & 100.0          & 99.6          & 99.8  & 25          \\
		TwoLeadECG                     & 2         & 82        & 23      & 1139   & 95.8 & 96.6          & 98.4           & \textbf{99.4} & 92.0  & 36          \\
		TwoPatterns                    & 4         & 128       & 1000    & 4000   & 95.8 & 93.1           & 95.2           & \textbf{99.4} & 69.6  & 33          \\
		UWaveGestureLibraryAll         & 8         & 945       & 896     & 3582   & \textbf{94.9} & 94.8 & 94.2           & 68.0          & 76.6  & 44          \\
		UWaveGestureLibraryX           & 8         & 315       & 896     & 3582   & 80.2 & 77.9          & 80.6           & 80.4          & 69.4  & 42          \\
		UWaveGestureLibraryY           & 8         & 315       & 896     & 3582   & 71.5 & 69.5           & \textbf{73.7}  & 71.8          & 59.1  & 40          \\
		UWaveGestureLibraryZ           & 8         & 315       & 896     & 3582   & 74.4 & 72.2          & 74.7           & 73.7          & 63.8  & 42          \\
		Wafer                          & 2         & 152       & 1000    & 6164   & 99.4 & 99.3          & \textbf{100.0} & 99.6          & 98.1  & 24          \\
		Wine                           & 2         & 234       & 57      & 54     & 86.5 & 86.4           & \textbf{92.6}  & 52.4          & 79.4  & 39          \\
		WordSynonyms                   & 25        & 270       & 267     & 638    & \textbf{68.4} & 62.8  & 58.2           & 58.1          & 46.1  & 42          \\
		Worms                          & 5         & 900       & 181     & 77     & 65.6 & 59.9          & \textbf{71.9}  & 64.2          & 62.2  & 47          \\
		WormsTwoClass                  & 2         & 900       & 181     & 77     & 74.1 & 69.1          & \textbf{77.9}  & 73.6          & 70.6  & 50          \\
		Yoga                           & 2         & 426       & 300     & 3000   & 83.3 & 80.2          & 82.3           & 83.3          & 72.1  & 40          \\
		
		\bottomrule
	\end{tabular}
	\caption{A comparison between \textsc{\texttt{gendis}} and three other shapelet techniques on 85 datasets. For each dataset, we report the total number of classes, the length of the time series, the number of time series in the train and test set, the accuracy score on the test set achieved by \textsc{\texttt{gendis}} (using both an ensemble (Ens) and logistic regression (LR)), \textsc{\texttt{st}}, \textsc{\texttt{lts}} \& \textsc{\texttt{fs}}, and finally the average number of shapelets extracted by \textsc{\texttt{gendis}}. When a technique is statistically significant better than the others, according to a student t-test with a cutoff of $0.05$, it is marked as bold.}
	\label{tbl:gendis_all_2}
\end{table}

\section{Conclusion}\label{sec:conclusion}

In this study, an innovative technique, called \textsc{\texttt{gendis}}, was proposed to extract a collection of smaller subsequences, i.e. shapelets, from a time series dataset that are very informative in classifying each of the time series into categories. \textsc{\texttt{gendis}} searches for this set of shapelets through evolutionary computation, a paradigm mostly unexplored within the domain of time series classification, which offers several benefits:
\begin{itemize}
	\item evolutionary algorithms are gradient-free, allowing for an easy configuration of the optimization objective, which does not need to be differentiable
	\item only the maximum length of all shapelets has to be tuned, as opposed to the number of shapelets and a length of each shapelet, due to the fact that \textsc{\texttt{gendis}} evaluates entire sets of shapelets
	\item easy control over the runtime of the algorithm
	\item the possibility of discovering shapelets that not need to be a subsequence of the input time series
\end{itemize}
Moreover, the proposed technique has a computational complexity that is multiple orders of magnitude smaller ($\mathcal{O}(GPKNM^2)$ vs $\mathcal{O}(N^2M^4)$) than current state-of-the-art, \textsc{\texttt{st}}, while outperforming it in terms of predictive performance, with much smaller shapelet sets. \\

We demonstrate these benefits through intuitive experiments where it was shown that techniques that evaluate single candidates can perform subpar on imbalanced datasets and how sometimes the necessity arises to extract shapelets that are no subsequences of input time series to achieve good separation. In addition, we compare the efficiency of the different genetic operators on six different datasets and assess the algorithm's stability by comparing the output of two different runs on the same dataset. Moreover, we conducted an extensive comparison on a large amount of datasets to show that \textsc{\texttt{gendis}} is competitive to the current state-of-the-art while having a much lower computational complexity.

\section{Reproducibility \& code availability}\label{sec:reproducibility}

An implementation of \textsc{\texttt{gendis}} in Python 3 is available on GitHub\footnote{\url{https://github.com/IBCNServices/GENDIS}}. Moreover, code in order to perform the experiments to reproduce the results is included.

\section{Acknowledgements}\label{sec:acknowledgements}

G. Vandewiele is funded by a PhD SB fellow scholarship of Fonds Wetenschappelijk Onderzoek (FWO) (1S31417N). F. Ongenae is funded by a Bijzonder OnderzoeksFonds (BOF) grant from Ghent University.

\small

\bibliographystyle{apalike}
\bibliography{ecjsample}

\end{document}